\documentclass[letterpaper]{article} 
\usepackage{aaai23}  
\usepackage{times}  
\usepackage{helvet}  
\usepackage{courier}  
\usepackage[hyphens]{url}  
\usepackage{graphicx} 
\usepackage{natbib}  
\usepackage{caption} 
\usepackage{multirow}
\usepackage{comment}
\usepackage{amsmath}
\usepackage{amsfonts}
\usepackage{amssymb}
\usepackage{amsbsy}
\usepackage{xcolor}
\usepackage{mathtools}
\usepackage{algorithm}
\usepackage{algorithmic}
\usepackage{pdflscape}

\usepackage{newfloat}
\usepackage{listings}
\DeclareCaptionStyle{ruled}{labelfont=normalfont,labelsep=colon,strut=off} 
\floatstyle{ruled}
\newfloat{listing}{tb}{lst}{}
\floatname{listing}{Listing}
\title{Evaluating the reliability of automatically generated pedestrian and bicycle crash surrogates}
\author{
    Agnimitra Sengupta\textsuperscript{\rm 1}, 
    S. Ilgin Guler\textsuperscript{\rm 2}\equalcontrib,
    Vikash Gayah\textsuperscript{\rm 3}\equalcontrib,
    Shannon Warchol\textsuperscript{\rm 4}\equalcontrib\\
}
\affiliations{
    \textsuperscript{\rm 1,2,3}Department of Civil \& Environmental Engineering, The Pennsylvania State University\\
    University Park, PA 16802, USA\\
    \textsuperscript{\rm 4}Kittelson and Associates, Inc.,
    Portland, OR 97204, USA\\
    \textsuperscript{\rm 1}avs6897@psu.edu, \textsuperscript{\rm 2}sig123@psu.edu, \textsuperscript{\rm 3}vvg104@psu.edu, \textsuperscript{\rm 3}swarchol@kittelson.com
}

\usepackage{bibentry}

\begin{document}

\maketitle
\begin{abstract}
Vulnerable road users (VRUs), such as pedestrians and bicyclists, are at a higher risk of being involved in crashes with motor vehicles, and crashes involving VRUs also are more likely to result in severe injuries or fatalities. Signalized intersections are a major safety concern for VRUs due to their complex and dynamic nature, highlighting the need to understand how these road users interact with motor vehicles and deploy evidence-based countermeasures to improve safety performance. Crashes involving VRUs are relatively infrequent, making it difficult to understand the underlying contributing factors. An alternative is to identify and use conflicts between VRUs and motorized vehicles as a surrogate for safety performance. Automatically detecting these conflicts using a video-based systems is a crucial step in developing smart infrastructure to enhance VRU safety. However, further research is needed to enhance its reliability and accuracy. The Pennsylvania Department of Transportation (PennDOT) conducted a study using video-based event monitoring system to assess VRU and motor vehicle interactions at fifteen signalized intersections across Pennsylvania to improve VRU safety performance. 
This research builds on that study to assess the reliability of automatically generated surrogates in predicting confirmed conflicts without human supervision using advanced data-driven models. The surrogate data used for analysis include automatically collectable variables such as vehicular and VRU speeds, movements, post-encroachment time, in addition to manually collected variables like signal states, lighting, and weather conditions. The findings highlight the varying importance and impact of specific surrogates in predicting true conflicts, some being more informative than others. The differences between significant variables that help identify bicycle and pedestrian conflicts were also examined, revealing critical distinctions. The findings can assist transportation agencies to collect the right types of data to help prioritize infrastructure investments, such as bike lanes and crosswalks, and evaluate their effectiveness. 
\end{abstract}

\section{Introduction}
Bicyclists and pedestrians are among the most vulnerable road users (VRUs), with a high risk of being involved in crashes with motor vehicles. According to the National Highway Traffic Safety Administration (NHTSA) \cite{nhtsa_bike,nhtsa_ped}, 6,516 pedestrians and 938 bicyclists were involved in fatal crashes with motor vehicles in the United States in 2020, indicating a 3.9\% and 9\% increase from 2019, respectively. Furthermore, these fatalities constitute about 17\% and 2.4\% of the total fatal crashes in the US in 2020, respectively. 
Although the majority of these crashes occurred at non-intersections, a significant number occurred at intersections. Crashes involving VRUs at intersections are often attributed to factors such as inadequate infrastructure, driver inattention or error, and failure to yield. The causes of such crashes can be complex and multi-factorial with varying contributing factors depending on the specific context and location. 

Crash prediction models \cite{lord2010statistical} have been implemented in traffic safety research to identify the causal factors of increased crash frequency and develop effective countermeasures. Macro-level crash analyses examine spatially aggregated crashes and models to determine contributing large-scale factors. The results of such analyses have found signalized intersection density, length of sidewalks, volume of road users, population density, proportion of heavy vehicles, and vehicle miles traveled to be contributing factors to VRU crashes \cite{cai2016macro,cai2017macro,nashad2016joint,ukkusuri2011random,zhang2015investigating, wier2009area, nchrp2022}. 
Conversely, micro-level analyses identify location-specific features like speed limits, signal phasing, temporal variation of traffic volume through the day, lighting conditions, and demographics (age, gender, etc.) as the significant factors contributing to VRU safety at individual locations \cite{abdel2005analysis,eluru2008mixed, lee2005comprehensive, prati2017using}.

Despite the high risk of crashes involving VRUs, these crashes are relatively rare events \cite{gdot_report}. As a result, traditional crash-based methods may not provide sufficient data to identify and prioritize locations and treatments for improving their safety. 
To address these challenges and provide a more comprehensive assessment beyond crash analysis, the use of surrogate safety measures based on traffic conflict analysis has been proposed \cite{perkins1968traffic}. 
A traffic conflict is defined here as a situation between two or more roadway users in which a collision may occur if one or more of the roadway users do not change their path (e.g., by making an evasive maneuver) \cite{parker1989traffic}. 
Surrogate safety measures (SSM) are used to quantify the likelihood of a collision in terms of severity of an interaction by determining the spatial and temporal proximity of two or more road users involved in a conflict.
The most commonly used SSMs include time-to-collision (TTC) \cite{hayward1972near}, post-encroachment time (PET) \cite{allen1978analysis}, deceleration-to-safety time (DST) \cite{hupfer1997deceleration}, gap time (GT) \cite{vogel2002characterizes}, and time-to-accident (TA) \cite{liu2017modeling} in addition to behavioral factors like aggressive lane changing, speeding, red-light violations, and traffic characteristics like volume, speed, and delay. These measures can provide valuable insights into the interaction between vehicles and VRUs at intersections and other roadway segments and help identify safety-critical events. For a comprehensive review of the use of SSM in crash modeling, please refer to \cite{arun2021systematic,johnsson2018search}.

\begin{table*}[!]
\caption{Definition of commonly used surrogate safety measures}\label{tab:surrogate}
\begin{center}
\begin{tabular}{ll}
\hline
Surrogate measure & Definition \\ \hline
Time-to-collision (TTC) & \begin{tabular}[c]{@{}l@{}}Time until two road users collide if both continue\\at the same speeds and along the same paths \cite{hayward1972near}\end{tabular} \\ \hline

\begin{tabular}[c]{@{}l@{}}Deceleration-to-safety\\time (DST)\end{tabular} & \begin{tabular}[c]{@{}l@{}}Deceleration required for second road user to reach \\ conflict point no earlier than when first user leaves it \cite{hupfer1997deceleration}\end{tabular} \\ \hline

Post-encroachment time (PET) & \begin{tabular}[c]{@{}l@{}}Time between the departure of first user from a conflict \\point and arrival of another user at conflict point \cite{allen1978analysis}\end{tabular} \\ \hline

Gap time (GT) & \begin{tabular}[c]{@{}l@{}}Time between second user arriving at conflict point after first\\ user leaves when both continue at same speed and paths \cite{vogel2002characterizes}\end{tabular} \\ \hline

Time to accident (TA) & \begin{tabular}[c]{@{}l@{}}Time until accident occurs after an evasive action is taken\\ if both continued with changed speeds and directions\cite{liu2017modeling}\end{tabular} \\ \hline

Conflict speed (CS) & \begin{tabular}[c]{@{}l@{}} Speed of the road user taking evasive action at \\moment before the evasive action is taken \cite{bode2019emergence}\end{tabular} \\ \hline
\end{tabular}
\end{center}
\end{table*}

Surrogate measures of safety are traditionally collected by manually observing conflicts over a period of time \cite{perkins1968traffic,baker1972evaluation}. However, the process of manual data collection is expensive, and the results are subject to inter- and intra-observer variability, which can reduce the repeatability and consistency of the data \cite{glauz1985expected,ismail2009automated, migletz1985relationships}. Although, simulation models \cite{sayed1994simulation, persaud1995microscopic, mehmood2001simulation} can account for some of these limitations, they cannot fully capture the heterogeneous and unpredictable driver behavior in real-world traffic.
Automated video-camera analysis \cite{autey2012safety, ismail2009automated, ismail2010automated} has emerged as a promising alternative to address the challenges associated with collecting conflict data through field observers or simulation models. This approach offers a complementary solution to improve data collection reliability while providing more in-depth analysis \cite{autey2012safety,ismail2009automated, ismail2010automated}. For example, an automated video-based pedestrian-vehicle conflict monitoring system \cite{ismail2009automated, ismail2010automated} was used to automatically identify potential conflicts based on four different surrogate measures -- namely TTC, DST, GT, and PET -- which was further validated with manually labeled events. 
A combination of all four indicators proved moderately effective in identifying important events and traffic conflicts. However, the study relied on various manually defined detection conditions over the composite values of the surrogate measures to identify critical events. Similarly, an automated bicycle-vehicle conflict detection using TTC and proximity as the surrogate measures was proposed in \cite{sayed2013automated} to detect likelihoods of critical events and conflicts that showed good agreement with confirmed conflicts, reporting an average accuracy of 87\%. Although these studies have mainly focused on using various conflict indicators to predict critical events, they were not used to predict confirmed conflicts. 

The Pennsylvania Department of Transportation (PennDOT) recently conducted a study (\textit{Smart Intersection Multimodal Safety Countermeasure Study}) using video-based event monitoring and crash data to assess the interactions of VRUs, both pedestrian and bicycle, with motor vehicles at signalized intersections throughout Pennsylvania, aiming to deploy engineering countermeasures more effectively to enhance safety performance \cite{penndot_report}. Fifteen intersections across Pennsylvania were monitored by the study team for multiple days, where video analysis was utilized to measure PET between motor vehicles and VRUs. Additionally, traffic counts, road user types, land use characteristics, traffic signal-related information, and speeds were collected to better understand the interactions of road users. 
The primary objective of the PennDOT project was to identify critical events, which were defined as events with a PET of less than three seconds, and manually review them to obtain confirmed conflicts, which were then used as surrogates and compared to actual crashes. However, the video-based event monitoring technology could not be used to automatically detect confirmed conflicts, creating a research gap. 

In this study, we propose multiple data-driven techniques to automatically detect confirmed conflicts from surrogate data gathered from the video-based event monitoring system of PennDOT. 
This approach has the potential to significantly improve the accuracy and efficiency of conflict identification and enhance our understanding of roadway safety risks. 
By incorporating ML techniques into the automated event detection framework, we can effectively capture complex patterns and relationships in the data and develop more reliable and robust conflict detection models.




\section{Data description}\label{sec:data}
The PennDOT study involved collecting video data for a duration of one week using Miovision (https://miovision.com/) cameras at fifteen intersection locations. The data was processed using TrafxSAFE technology
(https://safety.transoftsolutions.com/trafxsafe/) to analyze and automatically extract various parameters such as traffic volume, speed, trajectory, pedestrian and bicycle activities, and scenario type that provided a detailed description of the traffic event. Additional data on road user type, movements, conflict speeds, and median speeds were recorded for events with reported PETs between VRUs and motor vehicles of less than 10 seconds. 


The remainder of this work only focuses on critical events (i.e., PET less than 3 seconds). However, a low PET alone does not necessarily signify a conflict. Low PET events can occur when both drivers and VRUs are aware of the situation and there are no real safety concerns.
Hence, a manual review was carried out for up to 100 critical events per intersection to determine if they constituted a conflict based on the proximity, evasive action, and awareness of the involved road users. The classification of events as confirmed conflicts was also informed by more subjective characteristics such as the degree of recklessness displayed by drivers and VRUs, non-verbal communication between those involved, the context and intersection features, and any misjudgments made by either road user. 
In some cases, the available number of critical events per intersection was lower, or certain events were removed due to duplication, leading to fewer than 100 critical events per intersection. After the manual review, a dataset consisting of 1470 critical events with 89 confirmed conflicts was created. Notably, the response variable, which is the presence of confirmed conflicts, is very low and hence the dataset is highly unbalanced, which is a common occurrence in traffic safety studies.

Furthermore, since pedestrians and bicyclists have distinct characteristics, behaviors, and exposure, it is crucial to create models that are specific to each VRU to understand their individual safety concerns. However, due to the disproportionately lower number of observations related to bicycles compared to pedestrians (291 vs. 1179), it may not be feasible to develop a model specifically for bicycles. To address this issue, the results obtained from a model that considers both pedestrians and bicycles will be compared with a model that solely focuses on pedestrians. Therefore, summary statistics for the combined data and pedestrian-only data are presented here. 

\begin{figure*}[]
  \centering
  \includegraphics[width=0.75\textwidth]{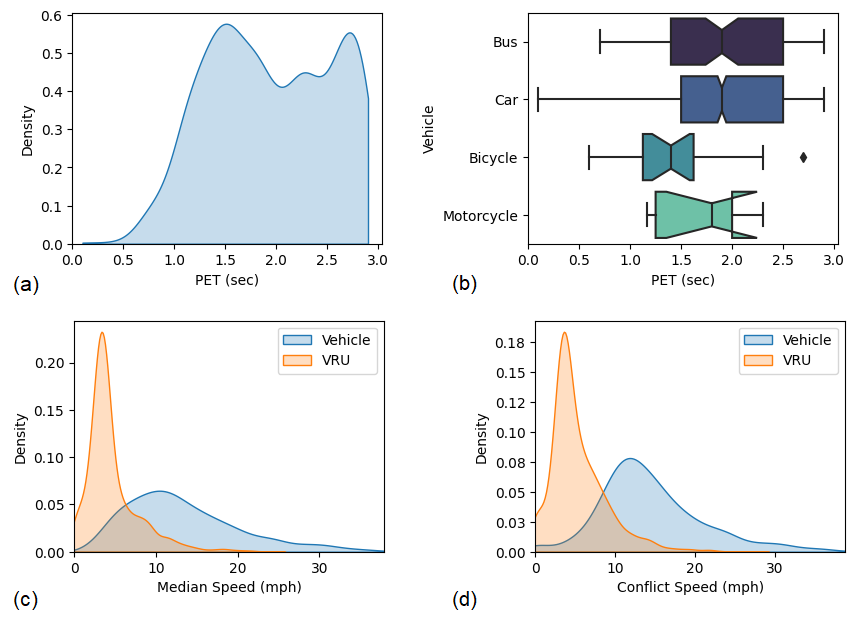}
  \caption{(a) PET distribution; (b) PET distribution by vehicle class; (c) median speeds and (d) conflict speeds for combined VRU dataset}\label{fig:continuous1}
\end{figure*}

\begin{figure*}[]
  \centering
  \includegraphics[width=0.75\textwidth]{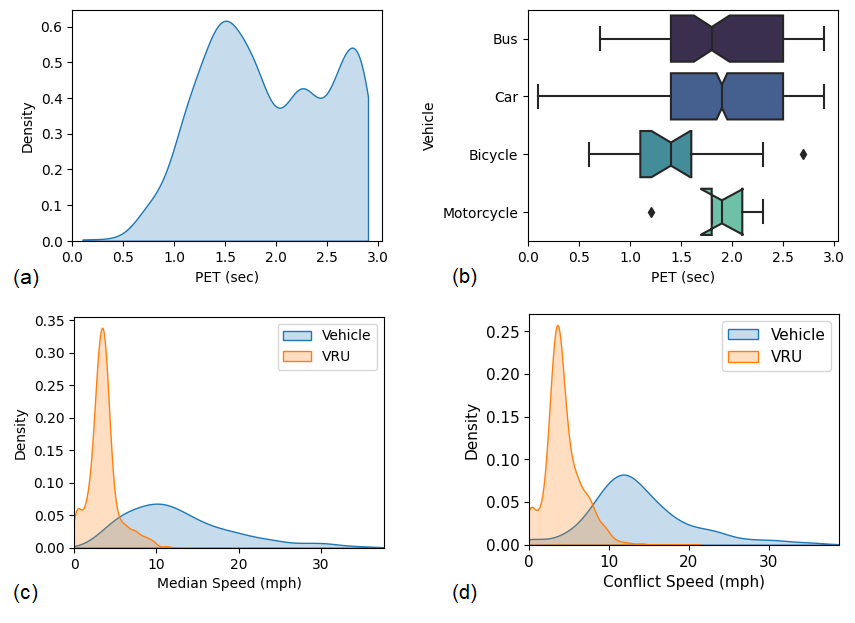}
  \caption{(a) PET distribution; (b) PET distribution by vehicle class; (c) median speeds and (d) conflict speeds for pedestrian-only dataset}\label{fig:continuous2}
\end{figure*}

Automatically extracted surrogates that were used as continuous input variables in the models include PET, median and conflict speeds of both vehicles and VRUs. 
These variables are recognized to have considerable impacts on the likelihood of traffic conflicts, and their integration can significantly enhance the performance of the models. 
The distribution plots for continuous variables in the combined data and pedestrian-only data are displayed in Figures~\ref{fig:continuous1} and \ref{fig:continuous2} respectively. 
Subplots (a) in both figures demonstrate that the PET distributions are bimodal with a slight skew towards lower values, indicating a decrease in the number of critical events with extremely low PET values (less than 1 second).
The comparison of PET distribution for both datasets indicates that the density of PETs greater than 2 seconds is higher for the combined data than the pedestrian-only data. This finding indicates that a significant number of critical events involve bicycles as the only VRU with PET greater than 2 seconds. 
Subplots (b) show the box-plots for the PET values grouped into vehicle classes, where `bicycle' refers to an interaction with a bicycle being treated as a vehicle and a pedestrian as a VRU. On average, the PET distribution for cars is greater than that of bicycles and motorcycles, while statistically similar to that of buses. However, the PET values of cars exhibit higher variation as compared to buses.
Subplots (c) and (d) in both figures demonstrate that the median and conflict speeds for VRUs are significantly lower than those for vehicles, as expected. 
The analysis of the combined dataset reveals that VRUs have a lower average median speed of 4.6 mph compared to vehicles with an average median speed of 13.3 mph. Additionally, VRUs exhibit a lower average conflict speed of 5.3 mph compared to vehicles with an average conflict speed of 14.4 mph. A similar trend is observed in the pedestrian-only dataset, where the average median speeds of VRUs and vehicles are 3.5 mph and 12.8 mph, respectively. Additionally, the average conflict speeds of VRUs and vehicles are 4.2 mph and 13.8 mph, respectively. 
These results suggest that the speeds of bicycles are higher than those of pedestrians. Moreover, the analysis indicates that the conflict speed is higher than the median speed of the road user, which underscores the importance of taking conflict speeds into consideration for safety analysis concerning VRUs. 

Table~\ref{tab:categorical} presents a summary of the statistical distribution of the categorical variables for the combined VRU and pedestrian-only datasets. To account for infrequent occurrences, some levels of categorical variables were grouped based on engineering judgment to reduce the number of levels and ensure proper estimation of the variable coefficients in the models.
For example, multiple levels of the variable `Vehicle type' -- such as articulated truck, box truck, single-unit truck, bus, pickup truck, and work-van -- were combined into a single category of heavy vehicles, referred to as `bus'.
The category `motorcycle' was not grouped with other vehicle types because of its unique operational differences, despite having only 7 observations. 
Additionally, `rain' and `snow' levels of the variable `Weather' were combined into a single category called `precipitation', due to their low frequency.
Similarly, the `VRU location' levels of `in crosswalk', `out of crosswalk', and `near crosswalk' were combined into a single level called `crosswalk'. 
To maintain consistency in reference to signal states, `do not walk' and `red ball' were merged into a single level called `red', while `green/yellow arrow and ball' were merged into a single level called `green' for both VRU and vehicle signal states. 

As mentioned earlier, the combined VRU dataset comprises nearly 80\% pedestrian-related data, while only 20\% corresponds to bicycles. In contrast, the pedestrian-only dataset focuses solely on pedestrian-related incidents. When examining the vehicles involved, we find that the majority of critical events involve cars, followed by bus-related incidents, while motorcycle-related incidents are the least frequent. 
It is worth noting that in cases where a pedestrian and a bicycle interact, the bicycle is categorized as a vehicle and the pedestrian as a VRU. However, such incidents are relatively rare. 
The distribution of road users arriving first at conflict points is quite similar in both datasets, with approximately 60\% of interactions occurring when cars arrive first, followed by pedestrians and buses. However, an interesting disparity in the pedestrian-only dataset is that less than 1\% of interactions involve bicycles arriving first.
The majority of the VRUs are located in the crosswalk, which follows from the fact that most VRUs in the dataset are pedestrians. Pedestrians are also located on the curb when waiting to use the crosswalk, with very few observations where pedestrians walking on the sidewalk involve a conflict. The travel lane is exclusively used by bicycles, where they share the right-of-way with other motor vehicles.
Upon examining the `VRU movement' variable, we observe that since the majority of the data pertains to pedestrians, movements on `crosswalk' accounts for 90\% of the data. The other movement categories in the combined dataset most likely correspond to bicycle movements. The vehicle movements are comparable in the two datasets, with slightly more observations belonging to right turns in the pedestrian-only dataset. Left turns have comparatively fewer observations. Additionally, the entering side of the intersection (nearside) with respect to VRU account account for a higher number of critical events as compared to the exiting side (farside). 
In both datasets, approximately 95\% of critical events mainly occur when the signal states of vehicles are green and remaining 5\% occur during the red state. In contrast, the distribution of signal states for VRUs is vastly different, with 60\% of the critical events occurring during the red state and 40\% during the green state. The signal states of VRUs in crosswalks are negatively associated with vehicle signal states. This implies that when the vehicle signal is green, the signal for the crosswalk is red, and any movement by a pedestrian or bicycle along the crosswalk would likely result in a conflict. However, critical events could also occur with vehicles that are turning left or right, when both the VRU and vehicle signal states are green.
The impact of weather and lighting conditions on the occurrence of critical events was also investigated. The distribution of weather conditions showed that most critical events were recorded during clear weather, followed by sunny weather. Additionally, about 80\% of the critical events were observed during daylight conditions, with few observations in the other conditions.  
However, it should be noted that an imbalance in data availability amongst levels of a categorical variable, i.e. some levels of the variable having a larger sample size compared to others, can often lead to inconsistent statistical inferences, which may be a limitation of the dataset.

\begin{table*}[!]
\caption{Summary statistics of categorical variables}\label{tab:categorical}
\begin{center}
\begin{tabular}{llll}
\hline
Variable & Levels & \begin{tabular}[c]{@{}l@{}}Combined (\%)\\$N$=1470\end{tabular} & \begin{tabular}[c]{@{}l@{}}Ped only (\%)\\$N$=1179\end{tabular} \\ \hline
\multirow{2}{*}{VRU} & Pedestrian & 80.13 &  100\\ 
& Bicycle & 19.87  &  -\\ \hline
 \multirow{4}{*}{Vehicle} & Bicycle & 1.22  & 1.44 \\ 
 & Bus & 7.96 &  8.23\\ 
 & Car & 90.34 & 89.91 \\ 
 & Motorcycle & 0.48 &  0.42\\ \hline
 
\multirow{5}{*}{Arrived first} & Bicycle & 7.69  &  0.59\\ 
 & Pedestrian & 25.65 &  31.98\\ 
 & Bus & 4.56 & 4.92 \\ 
 & Car & 61.90  &  62.26\\ 
 & Motorcycle & 0.20 & 0.25 \\ \hline

\multirow{4}{*}{VRU location} & Crosswalk & 70.20  &  78.63\\ 
 & Curb & 18.71 & 20.70 \\ 
 & Sidewalk & 0.54 &  0.68\\ 
 & Travel lane & 10.54 & -- \\ \hline

  \multirow{3}{*}{\begin{tabular}[c]{@{}l@{}}Vehicle\\movement\end{tabular}} & Through & 32.93  &  31.64\\ 
 & Left turn & 27.62 &  26.29\\ 
 & Right turn & 39.46 &  42.07\\ \hline
\multirow{2}{*}{Nearside} & Yes & 64.15 &  65.31\\ 
 & No & 35.85 & 34.69 \\ \hline
 \multirow{4}{*}{\begin{tabular}[c]{@{}l@{}}VRU\\movement\end{tabular}} & Crosswalk & 90.34  &  100\\ 
 & Through & 5.37 &  --\\ 
 & Left turn & 2.59 &  --\\ 
 & Right turn & 1.70 &  --\\ \hline
 
\multirow{2}{*}{Vehicle signal} & Green & 94.90 & 94.74 \\ 
 & Red & 5.10 & 5.26 \\ \hline
\multirow{2}{*}{VRU signal} & Green & 38.98 & 36.56 \\ 
 & Red & 61.02 & 63.44 \\ \hline
 
\multirow{4}{*}{Weather} & Clear & 50.75 & 49.19 \\ 
 & Sunny & 32.79 & 33.50 \\ 
 & Precipitation & 4.08 & 4.33 \\ 
 & Overcast & 12.38 &  12.98\\ \hline
\multirow{5}{*}{Lighting} & Daylight & 83.81 & 83.63 \\ 
 & Twilight & 1.97 &  1.61\\ 
 & Dark no streetlights & 0.61 &  0.76\\ 
 & Dark with streetlights & 8.91 &  9.84\\ 
 & Evening & 4.69 &  4.16\\ \hline

\end{tabular}
\end{center}
\end{table*}

\newpage
\section{Methodology}
In this section, we present the analytical framework used to predict pedestrian and bicycle confirmed conflicts using surrogate data generated from the automated event detection system. The goal is to use explanatory variables extracted from the video data to categorize low PET events (i..e, critical events) as a confirmed conflict. Two types of modeling approaches are considered for this binary categorization: 

\begin{enumerate}
    \item A statistical model, i.e., logistic regression
    \item Machine learning (ML) models, i.e., decision tree, random forest and extreme gradient boosting
\end{enumerate}
The ML methods require a hyperparameter optimization which is conducted using a Bayesian optimization framework. The models are evaluated using several performance measures. Additionally, the specific challenges associated with training an algorithm under class-imbalance due to the imbalance in available data is also considered.


\subsection{Logistic regression}
Logistic regression is a statistical technique used to model the probability of a binary outcome, such as success or failure. It uses a logistic function to describe the relationship between the independent variables ($\mathbf{x}$) and the probability of the dependent variable ($y$) taking a specific value, as shown below:

\begin{equation}
    p(y=1|\mathbf{x}) = \dfrac{1}{1+e^{-z}}
\end{equation}
\vspace{6pt}

where $z= \boldsymbol\beta\mathbf{x} = \beta_0 +\beta_1x_1+\cdots+\beta_px_p$, $\boldsymbol\beta$ is the vector of coefficients, and $\mathbf{x}$ is the vector of explanatory variables. The estimated coefficients $\boldsymbol\beta$ in a regression model indicate the influence of independent variables on the dependent variable. 

The odds ratio of a parameter reveals how a change in the independent variable affects the likelihood of a successful outcome. When a categorical variable ($x_j$) is included in the model, changing its indicator by one unit leads to an odds ratio of $e^{\beta_j}$. In contrast, for continuous variables, a $k$-fold change in the variable corresponds to an odds ratio of $k^{\beta_j}$. An odds ratio greater than 1 indicates an increase in the likelihood of a successful outcome, while a ratio less than 1 suggests a decrease. Therefore, positive coefficients in the model indicate an increased likelihood of success, while negative coefficients suggest a decreased likelihood.

The coefficients $\boldsymbol\beta$ can be estimated using the maximum likelihood estimation (MLE) method. Mathematically, for a binary classification problem with a training set of $N$ data points, where $y_i$ is the actual class label and $\mathbf{x}_i$ is the vector of input features for the $i^{th}$ data point, then the likelihood function can be formulated as:

\begin{equation}
    L(\boldsymbol\beta) = \prod_{i=1}^N p(y_i|\mathbf{x}_i,\boldsymbol\beta)^{y_i} \bigl(1- p(y_i|\mathbf{x}_i,\boldsymbol\beta)\bigr)^{1-y_i} \label{eqn:LL}
\end{equation}
\vspace{6pt}

Estimation of parameters is done through maximizing the natural logarithm of Equation~\ref{eqn:LL}, commonly known as log-likelihood. Although logistic regression is preferred due to its ease of implementation, superior interpretability and robustness to noise, it assumes a linear relationship between independent variables and probability of dependent variable on logit scale, which might not always hold true.

\subsection{Machine learning models}

Tree-based classification algorithms offer an interpretable and effective approach for addressing complex non-linear classification problems, making them well-suited for traffic safety-related studies where identifying critical variables influencing crashes and designing appropriate countermeasures are of utmost importance. All of the machine learning models used for this study -- decision trees, random forests, and extreme gradient boosting trees -- are tree-based models. 

\subsubsection{Decision tree}
A decision tree (DT) \cite{breiman2017classification} is constructed by recursively partitioning the input feature space into non-overlapping regions, where, the model makes a conditional split based on the values of the input features at each partitioning step. These splits lead to a complex tree that reflects the underlying data structure, with each terminal (or leaf) node containing a set of input data points with similar characteristics. Therefore, each split node in DT represents a test of a particular feature or attribute, and the branches represent the possible outcomes of the test. At each step of the tree-building process, the algorithm chooses the split based on `Gini' index for the available data at that node, which measures the total variance across all classes, see Equation \ref{eqn:gini}. 
\begin{equation}
    G= \sum_{c=1}^C \hat{p}_{mc}(1-\hat{p}_{mc})
    \label{eqn:gini}
\end{equation}
where $\hat{p}_{mc}$ represents the proportion of training observations in $m^{th}$ region that belong to $c^{th}$ class. 
It is evident that the Gini index takes on a small value if all $\hat{p}_{mc}$'s are close to zero or one. For this reason, the Gini index is referred to as a measure of node purity -- a small value indicates that a node contains predominantly observations from a single class. 
Once the tree is created, the mode of the response values of the  observations in the terminal node is used to classify a given input.

The process of recursively growing trees based on conditional if-else statements is likely to \textit{overfit}, leading to poor test performances. 
Therefore, we use techniques like pre-pruning to optimize on various tree hyperparameters, and post-pruning \cite{breiman2017classification} to obtain a sub-tree with lesser complexity, which has lower variance and better interpretation at the cost of bias. 

However pruning is not always sufficient to produce the most accurate and robust model. In such cases, ensemble models \cite{sagi2018ensemble} like random forests and gradient boosting trees are utilized to substantially improve the performance of trees. By combining inferences from multiple DTs, these models can reduce variance and increase the stability of the model to handle complex datasets with many features, thereby resulting in more accurate predictions.

\subsubsection{Random forest}
Random Forest (RF) is a ML algorithm that employs the bootstrap aggregation or bagging technique to generate multiple DTs. Each DT is built using bootstrap samples of the data by randomly selecting observations from the original training dataset. RF combines the decision of these multiple DTs to make more accurate decisions, i.e., the final prediction is determined by majority vote among the predictions of all trees in the forest. 

Although bootstrapped data can reduce model variance and improve test accuracy, it often generates DTs with highly correlated predictions. This correlation can compromise the effectiveness of the predictions since averaging correlated quantities can lead to biases in the prediction. 
To mitigate this problem, RF introduces an additional step in the creating of the individual DTs. At each split in a DT, a subset of predictors are randomly selected for consideration. This subset selection technique forces each DT to only use one of those predictors at each split, effectively decorrelating the trees and improving performance.

\subsubsection{Extreme gradient boosting}
Gradient boosting (GB) is an ensemble learning technique that utilizes a sequence of weak predictive models, each trained on the residuals of its predecessor, to construct a stronger learner.
In particular, gradient boosted decision tree (GBDT) model fits DTs sequentially, and at each step aims to learn from the errors at the previous steps by minimizing a loss function using a gradient descent procedure. Hence, each DT that is built is expected to have smaller errors than the previous one. Although GB is an efficient algorithm, it is limited by its computational complexity and can easily overfit the data. Extreme gradient boosting (XGBoost) \cite{chen2016xgboost}, which is based on the gradient boosting framework uses a second-order Taylor approximation on the loss function to capture additional information about the curvature of the loss function, and regularization parameters to prevent overfitting. 

\subsection{Learning with imbalanced data}
Training classification algorithms on imbalanced datasets can introduce unique challenges to the learning problem. 
A major concern with imbalanced data is the potential for it to severely compromise the performance of standard algorithms, rendering them ineffective.
Conventional ML algorithms assume balanced class distributions and equal misclassification costs. However, in complex imbalanced datasets, models tend to be biased towards the majority class, resulting in sub-optimal performance on the minority class. 
To address the effect of imbalanced datasets, we use techniques like:
\begin{enumerate}
    \item Cost-sensitive learning, which assigns different costs of misclassification to the individual data examples. Here, we assign class weights equal to inverse of their distributions, such that the minority class receives a higher cost, while the majority class is assigned a smaller misclassification cost.
    \item Data augmentation, where we balance the data distribution by synthetically generating samples belonging to minority class using a technique known as Synthetic Minority Oversampling Technique for Nominal and Continuous (SMOTE-NC) \cite{chawla2002smote}.
\end{enumerate}

\subsection{Performance measures}
\begin{figure}[!ht]
  \centering
  \includegraphics[width=.4\textwidth]{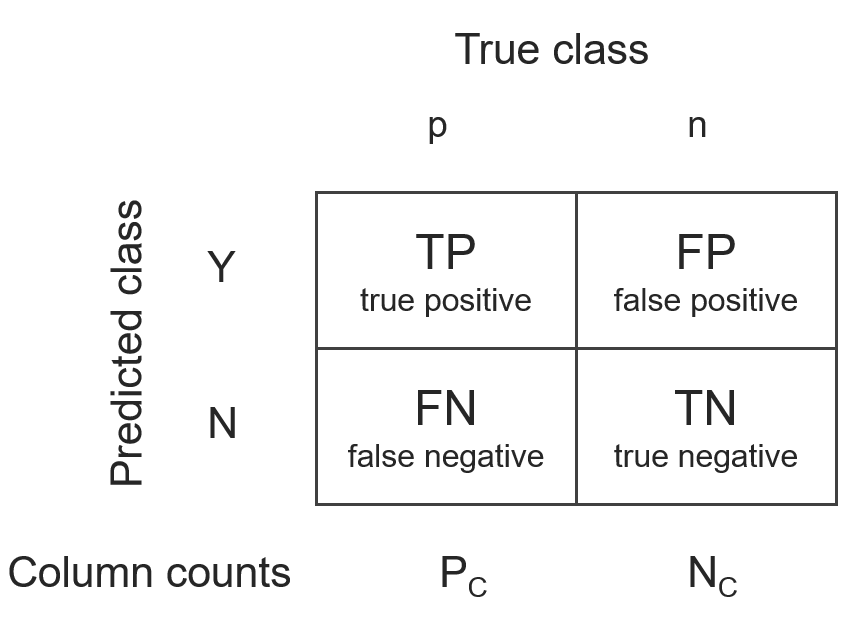}
  \caption{Confusion matrix for binary classification}\label{fig:confusion}
\end{figure}

In binary classification, the performance of an algorithm is typically assessed using a confusion matrix, as depicted in Figure~\ref{fig:confusion}. The matrix comprises of four categories, where true positive ($TP$) and true negative ($TN$) indicate correctly classified positive and negative class examples, and false positive ($FP$) and false negative ($FN$) represent incorrectly classified positive and negative class examples, respectively. When dealing with a balanced dataset and equivalent misclassification costs, accuracy is appropriate for evaluating the algorithm's overall effectiveness.

\begin{equation}
    \textit{Accuracy} = \dfrac{TP+TN}{P_C+N_C}\label{eqn:acc}
\end{equation}
\vspace{6pt}

However, accuracy can be impacted by changes in the data distribution and is unreliable when dealing with imbalanced datasets. In such cases, the algorithm may exhibit a bias toward the majority class and ignore misclassifications in the minority class. In the context of this study, the positive class indicating the presence of confirmed conflicts is represented by the minority class, whereas the negative class denoting the absence of confirmed conflicts is represented by the majority class.
To mitigate this limitation, precision, recall, and $F_1$ score are commonly used as defined below. These metrics take into account the classifier's performance on both positive and negative classes, resulting in a more comprehensive assessment of the algorithm's effectiveness.

\begin{equation}
    \textit{Precision} = \dfrac{TP}{TP+FP}
\end{equation}
\begin{equation}
    \textit{Recall} = \dfrac{TP}{TP+FN}
\end{equation}

\begin{equation}
    F_1 = \dfrac{\textit{Recall}\times\textit{Precision}}{\textit{Recall}+\textit{Precision}}
\end{equation}
\vspace{6pt}

Logistic regression and decision trees produce continuous numerical values that represent the confidence or probability of an instance belonging to a predicted class. To convert these probabilities into class labels, it is necessary to establish an appropriate threshold, above which the probability values are assigned to the positive class and below which they are assigned to the negative class. The choice of threshold can have a significant impact on the classifier's performance. To evaluate the binary classifier's performance across a range of thresholds, two graphical tools, namely the Receiver Operating Characteristics (ROC) curve and Precision-Recall (PR) curve, are utilized.

The ROC curve is generated by plotting the true positive rate, $TP_{rate}= {TP}/{P_C}$, against the false positive rate, $FP_{rate}= {FP}/{N_C}$, for different probability thresholds produced by the classifier. In contrast to the ROC curve, the Precision-Recall (PR) curve plots precision versus recall for a range of probability thresholds. The performance is evaluated by computing the AUC, or Area Under the Curve. The AUC value, which ranges from 0 to 1, serves as an overall measure of the classifier's performance, with a higher AUC value indicating better classification performance.

\section{Results and discussion}
In this section, we present a summary of the findings from the classification models, with a focus on their ability to accurately identify confirmed conflicts using crash surrogate data. 
To this end, we present the results of our models trained on a combination of pedestrian and bicycle data, followed by results on pedestrian-only data. We also evaluate the performance of these models on imbalanced data without and with cost-sensitive learning, as well as on a synthetically balanced dataset. Notably, models are trained and tested on 80\% and 20\% splits of the dataset. 
This allows us to assess the impact of imbalanced data on model performance and compare the effectiveness of different training approaches.

\subsection{Combined VRU model}
\subsubsection{Logistic regression}
The estimation results of the logistic regression model applied to the imbalanced dataset is presented in Table~\ref{tab:logit_imb}. The table provides information about the variable coefficients, their respective $p$-values indicating the level of significance, and the odds ratio of identifying confirmed conflicts. Among all candidate predictor variables, only five were statistically significant. The model's moderate fit to the observed data is reflected in the McFadden pseudo-$R^2$ value of 0.142. Notably, only certain levels of categorical variables were found to be statistically significant, resulting in the inclusion of only the significant levels in the model. 

To interpret the results of the logistic regression model, a critical event that was not a conflict was chosen as the baseline category for the target variable. The estimated coefficients were then interpreted based on the probability of a critical event being a confirmed conflict. 
The impacts of significant variables are discussed below:
\begin{itemize}
    \item Vehicle movement: The negative coefficient and odds ratio of less than 1 for the variable `through' movement suggests a weaker association of this movement with confirmed conflicts compared to the left turns (baseline). A critical event that happens during a left-turn movement is more likely to be a true conflict since for these movements the vehicles need to consider both the presence of other cars and VRUs which can lead to the omission of paying attention to the VRUs.
    
    \item VRU signal:   
    The estimated coefficients and odds ratio indicate a higher likelihood of a critical event involving pedestrians and bicycles being a conflict when the signal state for VRU is red. This could be due to VRUs violating the signal, resulting in conflicts with vehicles in motion and hence increasing the likelihood of an actual conflict.
    
    \item Proximity: Proximity refers to the distance between a vehicle and VRU involved in a critical event. A higher separation (or low proximity) would reduce the likelihood of a critical event being a conflict, which is expected since near misses are safety-critical events. 

     \item Post encroachment time: 
    PET is the time interval between the departure of one road user from a point on a roadway and the arrival of another road user at the same point. A lower PET value implies a shorter time interval between road users occupying the same space, which intuitively increases the likelihood of a confirmed conflict. The estimated negative coefficient and odds ratio of 0.35 for PET in the model provides support for this expectation, suggesting that lower PET values are more likely to be associated with confirmed conflicts.
 
    \item VRU conflict speeds: The higher the conflict speed of a VRU (i.e., the spot speed of the VRU at the moment at which the conflict is observed), the more likely a critical event is to be a confirmed conflict. This could be indicative of a reduced reaction time to respond to the VRU. 
    However, it is worth noting that the estimated coefficients for median speeds (i.e., the mean travel speed over the trajectory observable from the camera view) were found to be insignificant in the logistic regression, which may also be indicative of a VRU quickly accelerating to avoid a crash.
    
\end{itemize}

\begin{table*}[h]
\caption{Estimation results of logistic regression on imbalanced combined VRU dataset}\label{tab:logit_imb}
\begin{center}
\begin{tabular}{lllll}
\hline
Variable & Coefficient & Std. error & Odds ratio & $p>|z|$ \\ \hline
Intercept & -1.793 & 0.814 & 0.166 & 0.028 \\ 
Veh movement [though] & -1.132 & 0.373 & 0.322 & 0.002 \\ 
VRU signal [red] & 1.185 & 0.508 & 3.270 & 0.020 \\ 
Proximity [low] & -1.277 & 0.456 & 0.279 & 0.005 \\ 
PET & -1.042 & 0.369 & 0.486 & 0.005 \\
VRU conflict speed & 0.163 & 0.047 & 1.120 & 0.001 \\ \hline
McFadden $R^2$ & \multicolumn{4}{c}{0.142} \\ \hline
\end{tabular}
\end{center}
\end{table*}

\begin{figure*}[h]
  \centering
  \includegraphics[width=1\textwidth]{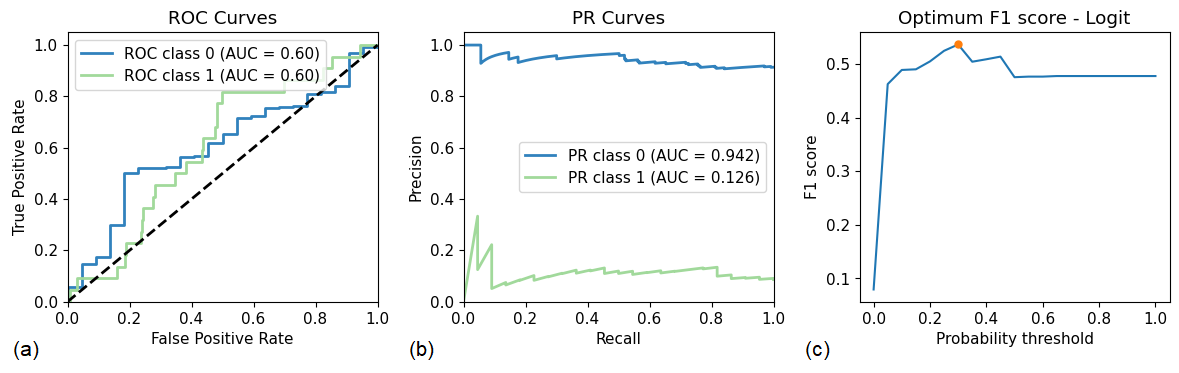}
  \caption{(a) ROC curves, (b) PR curves and (c) $F_1$ score versus probability thresholds for logistic regression on imbalanced combined VRU dataset}\label{fig:log_imb}
\end{figure*}

The fitted logistic regression model was used to predict the probabilities of observations belonging to the majority and minority classes in the test data. To evaluate the model's performance, Receiver Operating Characteristics (ROC) and Precision-Recall (PR) curves were employed as shown in Figure~\ref{fig:log_imb}, and both were compared for each class. The macro-average which refers to the mean performance across all classes is used to compare the overall classifier performances. Recall that a larger AUC for the ROC and PR curves indicates a better model performance. The ROC curve indicated that the logistic regression model achieved an AUC score of 0.60 on the imbalanced dataset. Furthermore, the AUC score of the PR curve demonstrated that the model's predictive ability varied for the two classes, suggesting that it may not be equally effective at predicting both classes. These findings imply that the model may have limited capability to learn discriminatory features of the minority class due to its limited examples, resulting in compromised predictive ability.

In order to assign class labels to our predictions, we needed to establish a threshold. However, the standard threshold of 0.50, typically used for balanced datasets or datasets with equal misclassification costs, was not suitable for our imbalanced dataset. Instead, we aimed to optimize the macro-average $F_1$ score for both classes as both classes were equally important, and thus determined an optimal probability threshold. 
The results, presented in Figure~\ref{fig:log_imb}(c), demonstrate the relationship between the macro-average $F_1$-score and various probability thresholds. We observed that a probability threshold of 0.30 yielded the highest $F_1$ score of 0.54 for our imbalanced dataset, and was therefore selected as the optimal threshold for our classification task.

Table~\ref{tab:class_logit_imb} presents a comparison of the logistic regression model's classification performance using the 0.50 threshold and the optimal threshold of 0.3. The results reveal a significant improvement in the model's classification performance when the optimal threshold was used, particularly for the minority class in the imbalanced dataset. Specifically, the macro $F_1$ score increased from 0.48 to 0.54. It is worth noting that the model's performance on the minority class could have been further enhanced by optimizing for the $F_1$ score of the minority class. However, such optimization would have increased the false positives in the dataset and led to sub-optimal overall performance. Therefore, we chose to optimize for the macro $F_1$ score, considering the importance of both classes. This observation underscores the importance of selecting an appropriate threshold for imbalanced datasets, as it can significantly affect the classification model's performance.

\begin{table*}[h]
\caption{Classification performance of logistic regression on imbalanced combined VRU dataset}\label{tab:class_logit_imb}
\begin{center}
\begin{tabular}{llllll}
\hline
Threshold & Conflict & Precision & Recall & $F_1$ score & Macro $F_1$\\ \hline
\multirow{2}{*}{0.50} & 0 & 0.91 & 0.99 & 0.95 & \multirow{2}{*}{0.48}\\ \cline{2-5}
 & 1 & 0.25 & 0.00 & 0.00 &    \\ \hline
\multirow{2}{*}{0.30} & 0 & 0.92 & 0.97 & 0.94 & \multirow{2}{*}{0.54}\\ \cline{2-5}
 & 1 & 0.22 & 0.09 & 0.13 &    \\ \hline
\end{tabular}
\end{center}
\end{table*}

\begin{figure*}[h]
  \centering
  \includegraphics[width=0.8\textwidth]{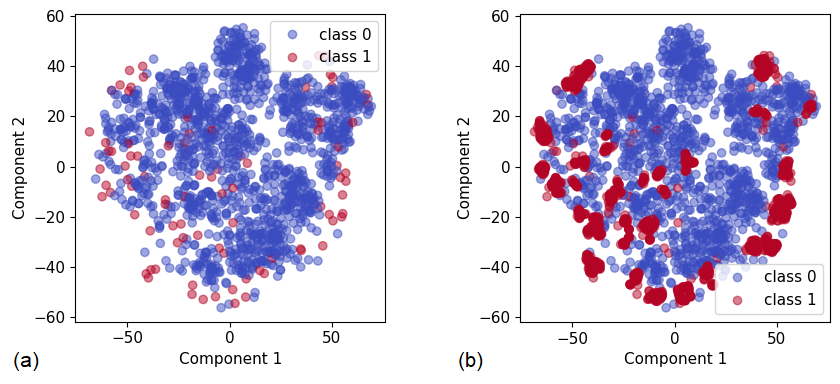}
  \caption{Low dimensional representation of (a) original dataset and (b) synthetically oversampled dataset using $t$-SNE}\label{fig:tsne}
\end{figure*}

To address the issue of class imbalance, we used cost-sensitive learning and employed a weighted logistic regression model with misclassification costs that were inversely proportional to the class distributions. Despite implementing this method, we did not observe any significant improvements in the model's performance. 
Furthermore, we employed a synthetically balanced dataset, where additional minority class data was synthetically generated using the SMOTE-NC technique. 
To visualize the low-dimensional representation of the original and synthetic datasets, we utilized $t$-distributed stochastic neighbor embedding ($t$-SNE) \cite{van2008}. This visualization revealed that the original and synthetic datasets are similar, as demonstrated by their close proximity in the $t$-SNE plot, shown in Figure~\ref{fig:tsne}. This suggests that the synthetic samples have a close representation of the underlying structure of the original dataset. 
It should be noted that although we used an oversampled dataset to enhance model training, we still evaluated model performance on real test data. Evaluating performance solely on synthetic data could result in inflated performance metrics that do not reflect actual performance.

The estimation results, shown in Table~\ref{tab:logit_bal}, indicate a significantly improved fit to the data compared to the previous models trained on imbalanced data. Additionally, we identified a few more significant variables in this model. Upon comparing these results to the previously trained model, we observed that both models shared many common variables. However, the balanced dataset model identified a few additional significant variables, which are described below:

\begin{itemize}
     \item VRU type: The coefficient and odds ratio for the pedestrian variable suggest that the probability of a critical event being a confirmed conflict is higher for vehicle-pedestrian events than for vehicle-bicyclist events. Several factors could contribute to this difference, including variations in road design, differences in behavior between pedestrians and bicyclists, or a lower number of observations for bicyclists in the dataset.
     
    \item Vehicle movement: Apart from through movement, which was already identified as significant, the estimation results identify right turn movement to be significant in identifying conflicts. While left turns are still considered the most safety-critical, the right turn movements are more likely to result in conflicts compared to through movements. This can be attributed to the presence of pedestrians along the cross street, which increases the probability of conflicts occurring.

    \item VRU movement: The only VRU movement identified as significant was right turn, indicating situations where bicycles are involved since pedestrian movements occur only along the crosswalk. The coefficient and odds ratio showed that this movement was less likely to be involved in a conflict, emphasizing the safety-critical events involving pedestrian through movements.

    \item Vehicle signal: The estimated coefficient and odds ratio for the vehicle signal being red indicate a lower likelihood of conflicts being associated with this variable. This is attributed to the fact that stationary vehicles are less likely to be involved in conflicts. This finding is also in line with our previous results, where we observed a positive coefficient for the red signal state for VRUs, indicating that conflicts involving pedestrians and bicycles are more likely to occur when vehicles are in motion.

    \item Vehicle Median speeds: In addition to VRU conflict speed and median speed, vehicle median speed was also observed to increase the likelihood of involvement in a conflict, which is a justifiable finding.

    \item VRU Median speed: A significant relationship was observed between both median speed and conflict speed of the VRU and the likelihood of a critical event being a confirmed conflict. The positive coefficient of the median speed variable indicates that higher median speeds of VRUs are associated with an increased likelihood of confirmed conflicts. However, the odds ratio of 1.092 suggests that the impact of median speed is smaller compared to that of conflict speed.
    
    \item Weather: The estimated coefficients and odds ratios indicate that overcast, precipitation and sunny weather have a lower likelihood of being involved in a confirmed conflict compared to the baseline (clear weather). However, we would expect that weather conditions like overcast or precipitation to adversely impact safety, and therefore have higher likelihood of being involved in a confirmed conflict. This discrepancy could be an artifact of the data since most observations were recorded in clear weather.
    
\end{itemize}

\begin{table*}[h]
\caption{Estimation results of logistic regression on balanced combined VRU dataset}\label{tab:logit_bal}
\begin{center}
\begin{tabular}{lllll}
\hline
Variable & Coefficient & Std. error & Odds ratio & $p>|z|$ \\ \hline
Intercept & -0.404 & 0.450 & 0.668 & 0.370 \\ 
Veh movement [right] & -0.580 & 0.159 & -0.560 & 0.000 \\
Veh movement [though] & -1.686 & 0.201 & 0.185 & 0.000 \\ 
VRU movement [right] & -3.385 & 1.125 & 0.034 & 0.003 \\ 
VRU type [pedestrian] & 1.334 & 0.278 & -3.797 & 0.000 \\ 
Veh signal [red] & -2.840 & 0.803 & 0.058 & 0.000 \\ 
Proximity [low] & -2.606 & 0.169 & 0.074 & 0.000 \\ 
PET & -0.962 & 0.121 & 0.513 & 0.000 \\ 
Veh conflict speed & 0.114 & 0.013 & 1.082 & 0.000 \\ 
VRU conflict speed & 0.160 & 0.034 & 1.117 & 0.000 \\ 
VRU Median speed & 0.177 & 0.039 & 1.130 & 0.000 \\ 
Weather [overcast] & -5.014 & 0.637 & 0.007 & 0.000 \\ 
Weather [precipitation] & -1.339 & 0.151 & 0.262 & 0.000 \\ 
Weather [sunny] & -2.941 & 0.593 & 0.053 & 0.000 \\ \hline
McFadden $R^2$ & \multicolumn{4}{c}{0.379} \\ \hline
\end{tabular}
\end{center}
\end{table*}

\begin{figure*}[h]
  \centering
  \includegraphics[width=1\textwidth]{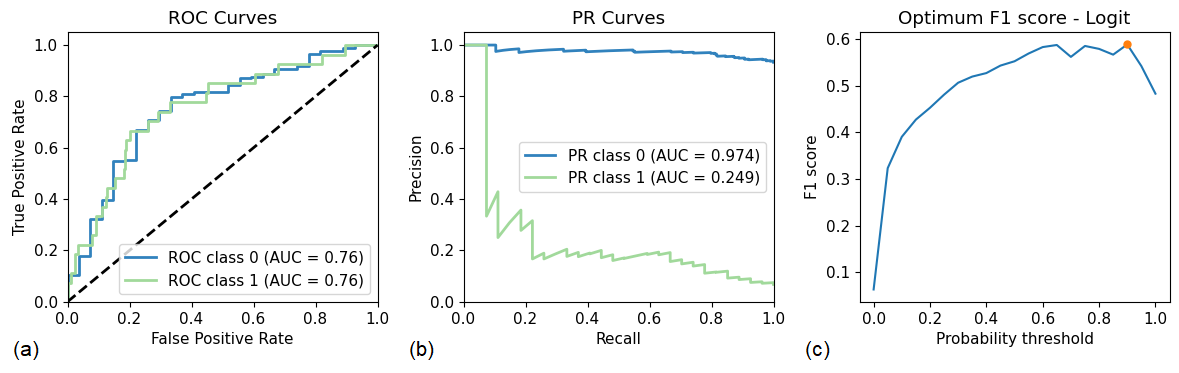}
  \caption{(a) ROC curves, (b) PR curves and (c) $F_1$ score versus probability thresholds for logistic regression on balanced combined VRU dataset}\label{fig:log_bal}
\end{figure*}

The logistic regression model exhibited a significantly improved performance in classifying the test set compared to the classification under imbalanced data conditions, both with and without cost-sensitive learning. Specifically, the macro average ROC AUC increased from 0.60 to 0.76, and the macro PR AUC improved from 0.534 to 0.612 when using the synthetically balanced dataset, see Figure~\ref{fig:log_bal}. These results suggest that the model was able to more accurately identify conflicts, indicating the effectiveness of the balanced data approach.
The results in Table~\ref{tab:class_logit_bal} indicate that using the optimum threshold provided a significant improvement in the classification performance of the model, the macro $F_1$ score improved from 0.55 to 0.59. Therefore, for the remainder of the paper only the results for balanced dataset are shown for brevity.
However, it is worth noting that there is a risk of overfitting on balanced data due to data augmentation based on a limited number of confirmed conflicts, leading to less variance in the synthetic samples.

\begin{table*}[h]
\caption{Classification performance of logistic regression on balanced combined VRU dataset}\label{tab:class_logit_bal}
\begin{center}
\begin{tabular}{llllll}
\hline
Threshold & Conflict & Precision & Recall & $F_1$ score & Macro $F_1$\\ \hline
\multirow{2}{*}{0.50} & 0 & 0.97 & 0.75 & 0.85 & \multirow{2}{*}{0.55} \\ \cline{2-5}
 & 1 & 0.16 & 0.67 & 0.26 &   \\ \hline
\multirow{2}{*}{0.90} & 0 & 0.94 & 0.97 & 0.95 & \multirow{2}{*}{0.59} \\ \cline{2-5}
 & 1 & 0.28 & 0.19 & 0.22 &   \\ \hline
\end{tabular}
\end{center}
\end{table*}



\subsubsection{Tree-based models}
To adequately prepare categorical variables for tree-based ML models, we utilize binary or one-hot encoding techniques to appropriately transform them. This encoding approach converts a categorical variable with $n$ levels into an $n$-dimensional variable, where each dimension represents the presence or absence of the corresponding level. However, when dealing with binary variables, we only keep one of the two indicator variables without loss of information to avoid redundancy. 
Following encoding of data, we utilized Bayesian optimization to select the optimal hyperparameters for models trained on the balanced dataset. The objective function optimized was the AUC of the PR curve (a.k.a average precision) for the minority class using a 3-fold cross-validation approach. The optimization results are presented in Table~\ref{tab:hyper_opt}.

\begin{table*}[h]
\caption{Hyperparameter optimization results on combined VRU dataset}\label{tab:hyper_opt}
\begin{center}
\begin{tabular}{llc}
\hline
Model & Hyperparameters & Objective function \\ \hline
Decision tree & \begin{tabular}[c]{@{}l@{}}Maximum depth of tree = 58\\ Minimum samples at leaf = 5\\ Minimum samples at split = 9\\ Pruning parameter = 0.0007\end{tabular} & 0.87\\ \hline
Random forest & \begin{tabular}[c]{@{}l@{}}Maximum depth of tree = 73\\ Minimum samples at leaf = 2\\ Minimum samples at split = 2\\ Number of estimators = 155\end{tabular} & 0.98\\ \hline
XGBoost & \begin{tabular}[c]{@{}l@{}}Subsample ratio of columns = 0.59\\ Learning rate = 0.20\\ Minimum loss at split = 0.48\\ Maximum depth of trees = 11\\ Minimum child weight = 2\end{tabular} & 0.95\\ \hline
\end{tabular}
\end{center}
\end{table*}

\begin{figure*}[]
  \centering
  \includegraphics[width=1\textwidth]{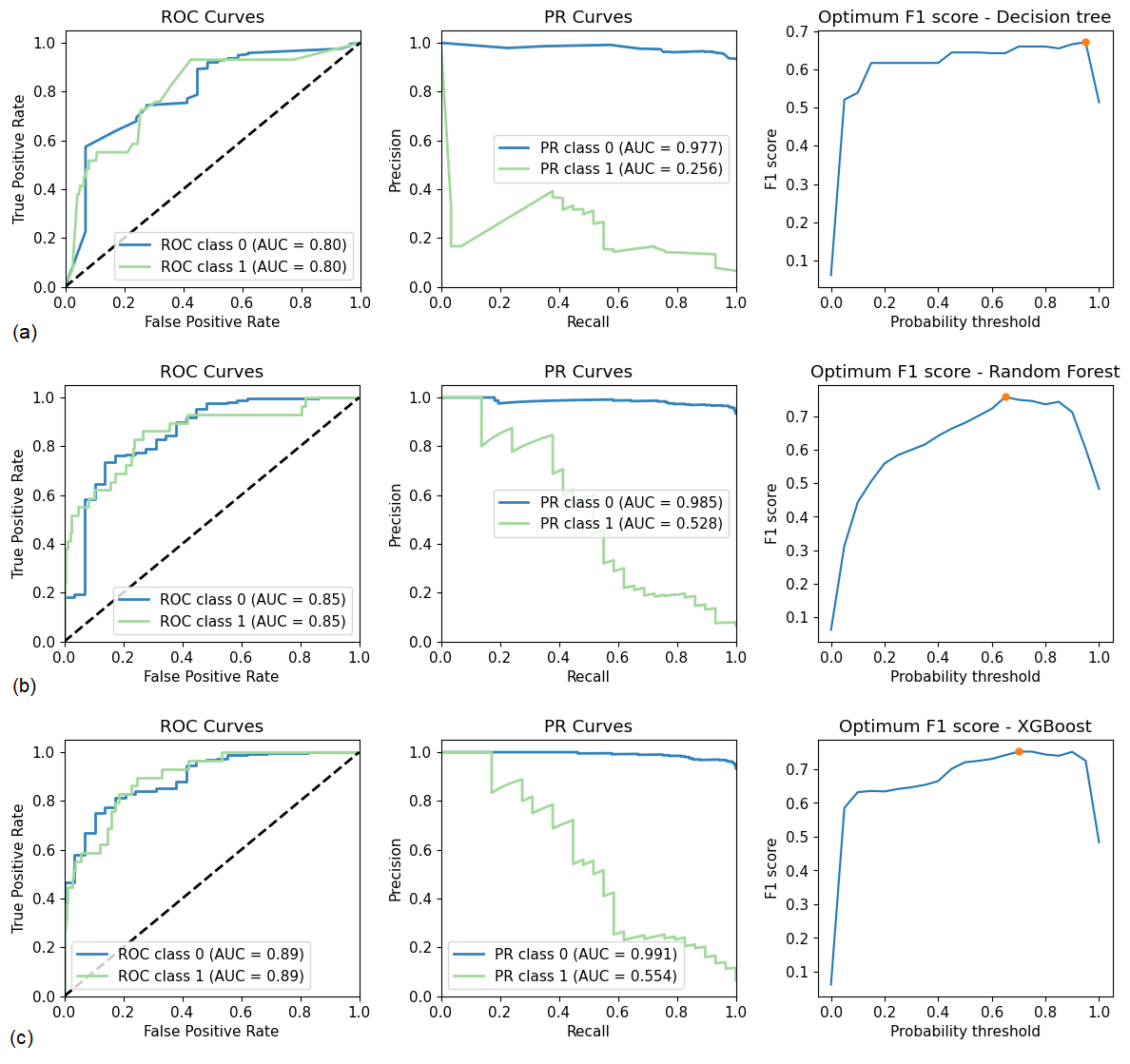}
  \caption{ROC curves, PR curves and $F_1$ score versus probability thresholds for (a) decision tree, (b) random forest, (c) XGBoost on balanced combined VRU dataset}\label{fig:tree_bal}
\end{figure*}

Using the optimized hyperparameters, tree-based models are trained and used to predict the class probabilities of observations on the test data. For all tree-based models, the ROC and PR curves for a balanced dataset are shown in Figure~\ref{fig:tree_bal}, and the  ROC AUC and PR AUC results are summarized in Table~\ref{tab:class_combined}. 
The macro F1 scores of all tree-based models indicate that these models clearly outperformed logistic regression for a balanced dataset. Further, optimizing the threshold was always found to increase the macro $F_1$ score. 



The outcomes revealed that in terms of macro $F_1$ score, the XGBoost model and RF model had similar performances, especially when the optimized threshold was used (macro $F_1$ of 0.75 and 0.76, respectively). The DT algorithm, while outperforming the logistic regression model, had the lowest macro $F_1$ even after threshold optimization (0.67) out of all of the tree-based models.

\begin{table*}[]
\caption{Classification performance comparison of tree-based models on balanced combined VRU dataset}\label{tab:class_combined}
\begin{center}
\begin{tabular}{llllllc}
\hline
Model & Threshold & Conflict & Precision & Recall & $F_1$ & Macro $F_1$\\ \hline
\multirow{4}{*}{Decision tree} & 
\multirow{2}{*}{0.50} & 0 & 0.97 & 0.89 & 0.93 & \multirow{2}{*}{0.64} \\ \cline{3-6}
&  & 1 & 0.27 & 0.55 & 0.36 &   \\ \cline{2-7} 
 & \multirow{2}{*}{0.95} & 0 & 0.96 & 0.95 & 0.95 & \multirow{2}{*}{0.67} \\ \cline{3-6}
& & 1 & 0.36 & 0.41 & 0.39 &     \\ \hline

\multirow{4}{*}{Random Forest} & 
\multirow{2}{*}{0.50} & 0 & 0.97 & 0.91 & 0.94 & \multirow{2}{*}{0.68} \\ \cline{3-6}
&  & 1 & 0.33 & 0.59 & 0.42 &   \\ \cline{2-7} 
 & \multirow{2}{*}{0.65} & 0 & 0.97 & 0.97 & 0.97 & \multirow{2}{*}{0.76} \\ \cline{3-6}
& & 1 & 0.58 & 0.52 & 0.55 &    \\ \hline

\multirow{4}{*}{XGBoost} & 
\multirow{2}{*}{0.50} & 0 & 0.97 & 0.94 & 0.96 & \multirow{2}{*}{0.72} \\ \cline{3-6}
&  & 1 & 0.41 & 0.59 & 0.49 &  \\ \cline{2-7} 
 & \multirow{2}{*}{0.65} & 0 & 0.97 & 0.97 & 0.97 & \multirow{2}{*}{0.75} \\ \cline{3-6}
& & 1 & 0.56 & 0.52 & 0.54 &    \\ \hline

\end{tabular}
\end{center}
\end{table*}

\subsubsection{Model interpretability}
\begin{landscape}
\begin{figure}[h]
  \centering
  \includegraphics[width=1\linewidth]{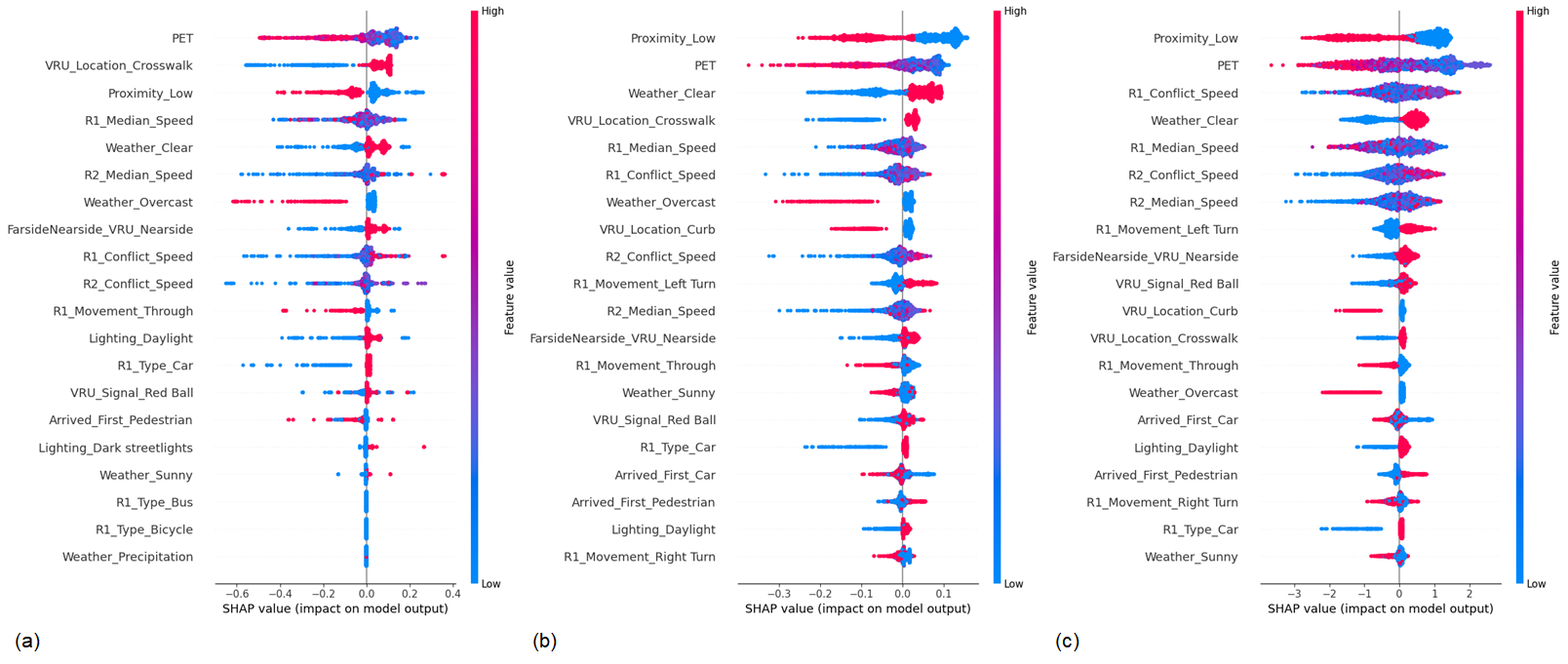}
  \caption{SHAP plots for (a) decision tree, (b) random forest, (c) XGBoost on balanced combined VRU dataset}\label{fig:shap_bal_combined}
\end{figure}
\end{landscape}

To evaluate the impact of individual features on the performance of a model, global feature importance is typically used. However, these metrics do not provide information regarding the specific contribution of variables to the output of individual observations. In this study, we employed Shapley additive explanations (SHAP) \cite{lundberg2017unified} values to analyze the contribution of each feature variable for every observation in driving the propensity of a critical event towards a conflict. We examined bee-swarm values for each ML classifier trained on balanced datasets. The results are presented in Figure~\ref{fig:shap_bal_combined}. Bee-swarm plots show the SHAP values for each feature as a scatter plot, with each point representing an instance in the dataset. For any feature, the position of points on the horizontal axis indicates the SHAP value for those instances, while the vertical dispersion is an indicator of the density of instances that share the same SHAP value. In addition, the color of each point in the plot represents its corresponding feature value, where red signifies a `high' value and blue signifies a `low' value. For continuous features, the threshold for determining `high' and `low' values is set at the maximum and minimum values of the feature, respectively. For binary variables, `high' and `low' values correspond to the presence and absence of the feature, respectively.

Comparing the SHAP values across classifier models, features related to the proximity, PET, and speeds of the vehicle and VRU consistently emerge as the most important across models. However, the feature values are heavily discounted in the DT model. Notably, features such as vehicle movement and VRU type, which are known to significantly affect the likelihood of a critical event resulting in a conflict, exhibit near-zero SHAP values in the DT model. 
This finding can be attributed to the limited number of observations in the dataset that have these specific feature values, as well as the tendency of DT models to overfit. 
In contrast, the RF and XGBoost models, which are ensemble models that leverage the collective inferences from multiple weak learners, have a reduced risk of overfitting, leading to a more comprehensive estimation of feature importance. In both models, variables such as signal states, vehicle and VRU movements, weather, and lighting conditions, alongside PET and conflict speeds, exhibit influence on the target variable. 
Similar observations were made regarding the impact of weather conditions, as in the case of logistic regression. However, specific weather conditions like precipitation and overcast weather seem to have a lower propensity for being involved in a conflict. This counter-intuitive effect could potentially be an artifact of the data since a majority of the observations were recorded in clear weather conditions.

This observation could be attributed to the dataset's characteristics, with most of the recorded observations taking place under clear weather and daylight conditions. 
It is important to note that balancing a dataset through synthetic approaches may alter the inherent distribution of variables, which can, in turn, affect feature importance. Therefore, it is crucial to carefully examine the impact of the balancing approach on the dataset and assess any changes in feature importance accordingly to ensure the reliability and validity of the results.

Moreover, we also observe that the impact of each identified independent variable on the target variable remains consistent across various models and learning scenarios. For example, we find that the SHAP values for the `Proximity' variable in all models consistently indicate that a high proximity value increases the likelihood of a confirmed conflict, while a low proximity value decreases it. Moreover, we observe that lower PET values consistently increase the likelihood of a critical event resulting in a conflict. Additionally, we find that as the conflict speed and median speed increase, the probability of a confirmed conflict also increases. These findings are in agreement with the estimation results of logistic regression, reinforcing their robustness and reliability.

\subsection{Pedestrian only model}
In this section, we highlight the key differences that have been observed for pedestrian-only model trained on balanced datasets. 

\subsubsection{Logistic regression}
The logistic regression analysis results for the pedestrian-only model on the balanced dataset are presented in Table~\ref{tab:logit_bal_ped}. The analysis suggests that the model fits the data well, as evidenced by the McFadden $R^2$ value of 0.363, although this value is slightly lower than what was observed in the combined VRU dataset.
Comparing the results of the pedestrian-only model with those of the combined VRU model, we observe some similarities as well as important differences. The variable `Nearside' is found to be significantly influential in increasing the likelihood of a critical event towards a conflict, which is expected because pedestrians positioned on the nearside of the intersection with respect to the vehicles are more likely to experience a conflict. Unlike the combined dataset, we found that conflict and median speeds are significant factors for both pedestrians and vehicles in the pedestrian-only dataset.

\begin{figure*}[h]
  \centering
  \includegraphics[width=1\textwidth]{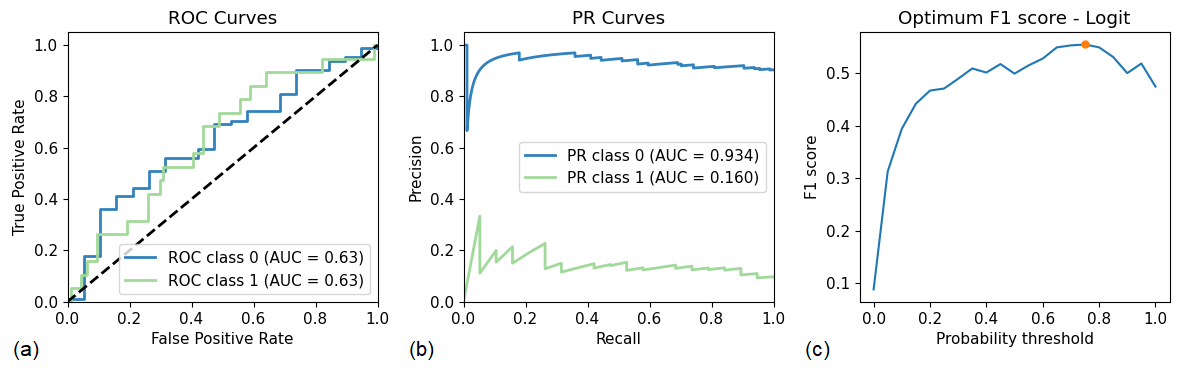}
  \caption{(a) ROC curves, (b) PR curves and (c) $F_1$ score versus probability thresholds for logistic regression on balanced pedestrian-only dataset}\label{fig:log_bal_ped}
\end{figure*}

After training the logistic regression model, we use it to predict the class probabilities for each observation pertaining to the test set and evaluate the classification performance using the AUC score for ROC and PR curves. The analysis shows that the model has slightly lower performance compared to the combined VRU model, as shown in Figure~\ref{fig:log_bal_ped}. 
To assign class labels to the observations, we select a probability threshold that maximizes the macro $F_1$ score for both classes. The optimal $F_1$ score in this case is 0.55, corresponding to a threshold of 0.75. This threshold provides a significant improvement over the macro $F_1$ score of 0.50 achieved using a threshold of 0.50. We provide a comparison of the classification performance using both thresholds in Table~\ref{tab:class_ped}.

\begin{table*}[!ht]
\caption{Estimation results of logistic regression on balanced pedestrian-only dataset}\label{tab:logit_bal_ped}
\begin{center}
\begin{tabular}{lllll}
\hline
Variable & Coefficient & Std. error & Odds ratio & $p>|z|$ \\ \hline
Intercept & -1.748 & 0.460 & 0.174 & 0.000 \\ 
Veh movement [though] & -2.430 & 0.289 & 0.088 & 0.000 \\ 
VRU signal [red] & 0.753 & 0.242 & 2.123 & 0.002 \\ 
Proximity [low] & -2.396 & 0.251 & 0.091 & 0.000 \\ 
PET & -1.111 & 0.184 & 0.463 & 0.000 \\ 
Nearside & 0.782 & 0.207 & 2.187 & 0.000 \\ 
VRU conflict speed & 0.285 & 0.049 & 1.219 & 0.000 \\ 
VRU Median speed & 0.228 & 0.055 & 1.172 & 0.000 \\ 
Veh conflict speed & 0.094 & 0.023 & 1.067 & 0.000 \\ 
Veh median speed & 0.056 & 0.026 & 1.040 & 0.031 \\ 
Weather [overcast] & -3.511 & 0.659 & 0.030 & 0.000 \\ 
Weather [precipitation] & -0.745 & 0.208 & 0.475 & 0.000 \\ 
Weather [sunny] & -3.623 & 1.218 & 0.027 & 0.003 \\ \hline
McFadden $R^2$ & \multicolumn{4}{c}{0.363} \\ \hline
\end{tabular}
\end{center}
\end{table*}

\subsubsection{Tree-based models}
\begin{figure*}[h]
  \centering
  \includegraphics[width=1\textwidth]{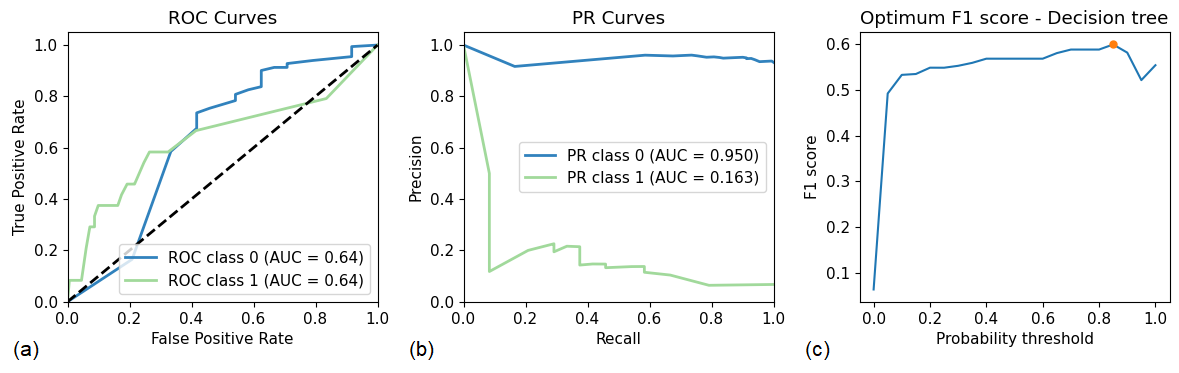}
  \caption{(a) ROC curves, (b) PR curves and (c) $F_1$ score versus probability thresholds for decision tree on balanced pedestrian-only dataset}\label{fig:dt_bal_ped}
\end{figure*}

Results of DT trained on pedestrian-only dataset indicates that the model's classification performance has slightly improved when compared to the logistic regression model, specifically in terms of ROC AUC and PR AUC. See Figure~\ref{fig:dt_bal_ped}. For example, the macro ROC AUC has increased from 0.63 to 0.64, while the macro PR AUC has improved from 0.547 to 0.556. Additionally, by optimizing the threshold, we have increased the macro $F_1$ score from 0.57 to 0.60, which is higher than that of the logistic regression model, as we see in Table~\ref{tab:class_ped}. 

However, DTs are known to overfit. To overcome the limitations of DT models, we conducted a performance evaluation of RF and XGBoost models on the test data. We found that these models outperformed both logistic regression and DT models, exhibiting macro AUC for ROC curves of 0.81 and 0.83 respectively, which is significantly higher than that of logistic regression and DT (0.63 and 0.64, respectively). Furthermore, there was substantial improvement in the macro AUC for PR curves, with values of 0.641 and 0.662 for RF and XGBoost, respectively, as compared to the DT model. This was largely attributed to the increase in precision for minority class predictions in both models.
We also observed that optimizing the threshold values for these models led to a significant improvement in their average performance for identifying confirmed conflicts. Specifically, we achieved macro $F_1$ scores of 0.65 and 0.69 for RF and XGBoost models, respectively, which are notably higher than the DT model's performance.


\begin{figure}[h]
  \centering
  \includegraphics[width=1\textwidth]{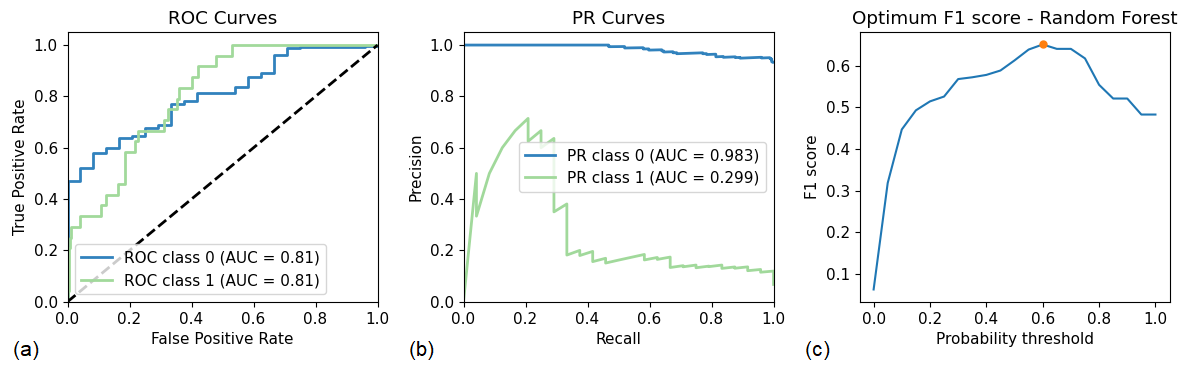}
  \caption{(a) ROC curves, (b) PR curves and (c) $F_1$ score versus probability thresholds for random forest on balanced pedestrian-only dataset}\label{fig:rf_bal_ped}
\end{figure}

\begin{figure}[h]
  \centering
  \includegraphics[width=1\textwidth]{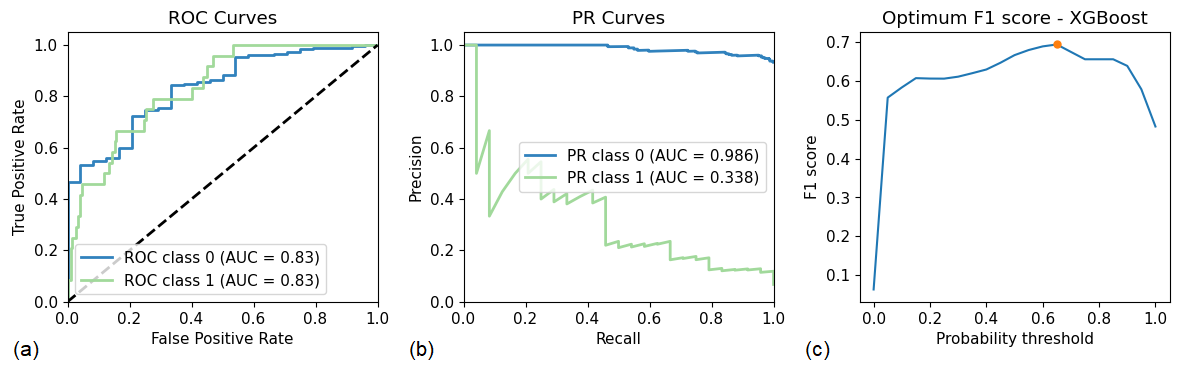}
  \caption{(a) ROC curves, (b) PR curves and (c) $F_1$ score versus probability thresholds for XGBoost on balanced pedestrian-only dataset}\label{fig:xg_bal_ped}
\end{figure}

\begin{table*}[h]
\caption{Classification performance comparison of XGBoost models}\label{tab:class_ped}
\begin{center}
\begin{tabular}{llllllc}
\hline
Model & Threshold & Conflict & Precision & Recall & $F_1$ & Macro $F_1$\\ \hline
\multirow{4}{*}{Logistic regression} & \multirow{2}{*}{0.50} & 0 & 0.91 & 0.75 & 0.82 & \multirow{2}{*}{0.50} \\ \cline{3-6}
& & 1 & 0.12 & 0.32 & 0.17 &   \\ \cline{2-7} 
 & \multirow{2}{*}{0.75} & 0 & 0.91 & 0.90 & 0.91 & \multirow{2}{*}{0.55} \\ \cline{3-6}
& & 1 & 0.19 & 0.21 & 0.20 &     \\ \hline

\multirow{4}{*}{Decision tree} & 
\multirow{2}{*}{0.50} & 0 & 0.95 & 0.86 & 0.91 & \multirow{2}{*}{0.57} \\ \cline{3-6}
&  & 1 & 0.17 & 0.38 & 0.23 &  \\ \cline{2-7} 
 &\multirow{2}{*}{0.85} & 0 & 0.95 & 0.90 & 0.93 & \multirow{2}{*}{0.60}\\ \cline{3-6}
& & 1 & 0.24 & 0.38 & 0.27 &    \\ \hline

\multirow{4}{*}{Random forest} & 
\multirow{2}{*}{0.50} & 0 & 0.95 & 0.93 & 0.94 & \multirow{2}{*}{0.61} \\ \cline{3-6}
&  & 1 & 0.25 & 0.33 & 0.29 &   \\ \cline{2-7} 
 & \multirow{2}{*}{0.60} & 0 & 0.95 & 0.97 & 0.96 & \multirow{2}{*}{0.65}\\ \cline{3-6}
& & 1 & 0.41 & 0.29 & 0.34 &   \\ \hline

\multirow{4}{*}{XGBoost} & 
\multirow{2}{*}{0.50} & 0 & 0.96 & 0.93 & 0.95 & \multirow{2}{*}{0.67}\\ \cline{3-6}
&  & 1 & 0.33 & 0.46 & 0.39 &  \\ \cline{2-7} 
 & \multirow{2}{*}{0.65} & 0 & 0.96 & 0.95 & 0.96 & \multirow{2}{*}{0.69}\\ \cline{3-6}
& & 1 & 0.41 & 0.46 & 0.43 & \\ \hline

\end{tabular}
\end{center}
\end{table*}

\subsubsection{Model interpretability}
Upon analyzing the effects of variables using SHAP plots, we observed both similarities and distinct differences in the pedestrian-only model and the combined VRU model. The DT model displayed a lack of importance for many features, possibly due to overfitting. However, the ensemble models provided a more comprehensive understanding of variable importance.
Similar to our previous findings in the combined VRU dataset, we observed that variables such as proximity, median and conflict speeds, and PETs had similar effects on the propensity of a critical event toward a conflict in the pedestrian-only dataset. For example, higher proximity and low PET values were associated with an increase in the likelihood of conflicts, as indicated by positive SHAP values.
A critical difference we observed was that through movements of vehicles were found to be significant in the pedestrian-only dataset while showing lower SHAP importance in the combined VRU dataset. This could be due to the fact that most pedestrian conflicts are associated with the through movement of vehicles while pedestrian movement is along the crosswalk.
Similar to our previous findings, we found that weather and lighting were significant variables with high SHAP values. However, despite their significance, these variables are not expected to increase the likelihood of conflicts. This could be due to potential misinterpretations arising from the inherent distribution of the data. It is noteworthy that these findings are consistent with the results obtained from the logistic regression estimation.


\begin{landscape}
  \begin{figure}[h]
  \centering
  \includegraphics[width=1\linewidth]{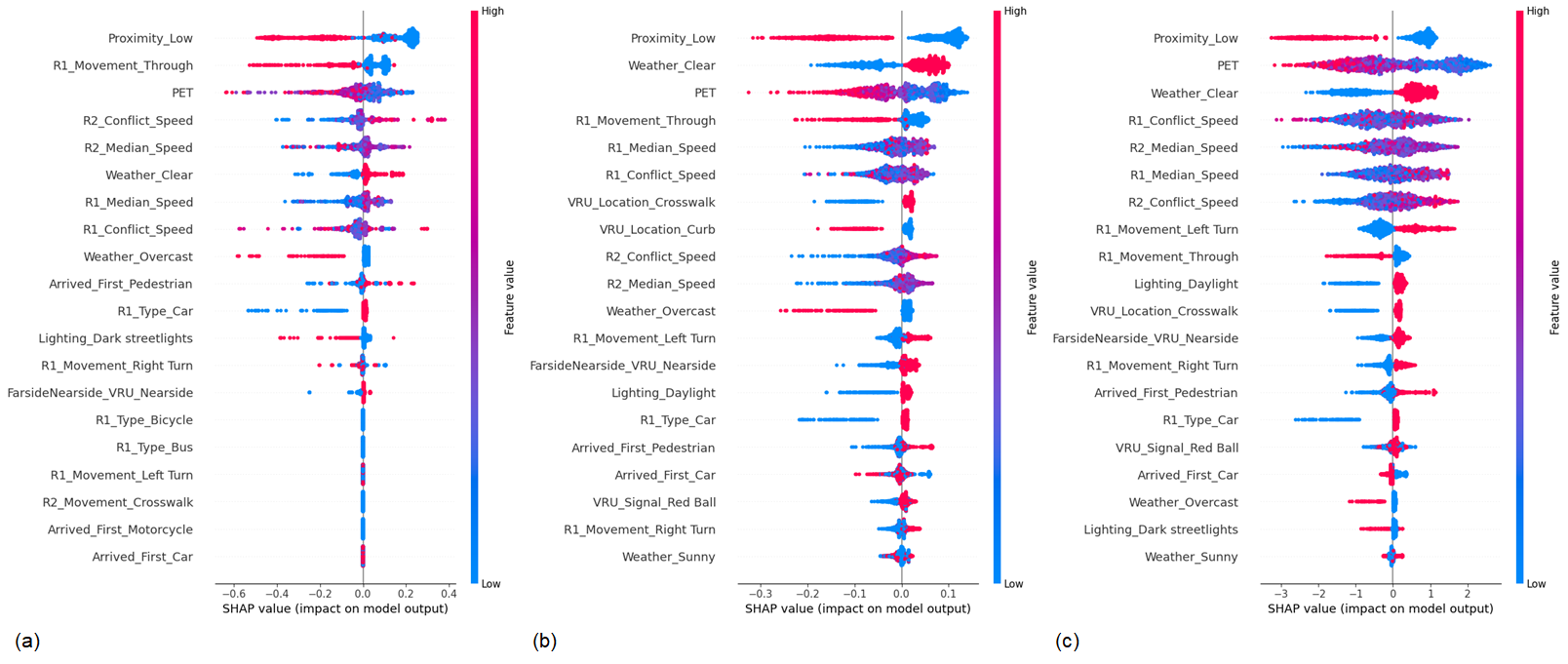}
  \caption{SHAPley plots for (a) decision tree, (b) random forest, (c) XGBoost on balanced pedestrian-only dataset}\label{fig:logit_bal_auc_ped}
\end{figure}  
\end{landscape}

\section{Conclusions}
Pedestrian and bicycle (VRU) safety is a critical issue given the vulnerability of these road users to fatalities. However, since these crashes are relatively infrequent, surrogate measures based on frequently observed critical events are often utilized to evaluate VRU safety \cite{perkins1968traffic,svensson2006estimating}. 
In this study, we present and compare multiple data-driven frameworks to automatically predict confirmed conflicts from surrogates obtained through PennDOT's video-based event monitoring system. 
Unfortunately, these models trained on an imbalanced dataset display bias towards the majority class (which, in this case, is the absence of conflicts). This can lead to sub-optimal performances in the minority class (presence of conflicts), which is demonstrated in this paper.
State-of-the-art approaches, such as data augmentation and cost-sensitive learning techniques, have been utilized in the study for addressing imbalanced data in model training and have been shown to significantly enhance model performance.

Comparing the model performance of different approaches shows that logistic regression and decision tree (DT) models perform similarly in detecting conflicts for both imbalanced and balanced datasets. However, DT models tend to overfit the data, which is addressed by utilizing cost-complexity pruning. In contrast, ensemble learning techniques such as random forest (RF) and extreme gradient boosting (XGBoost), which utilize bagging and gradient boosting, respectively, have proven to be effective in handling high-dimensional data with a lower risk of overfitting. Therefore, RF and XGBoost offer a dependable interpretation of the impact of various surrogates on conflict identification. Furthermore, these modeling techniques highlight the varying importance and impact of specific surrogates in predicting true conflicts, some being more informative than others.
However, as pedestrians and bicycles exhibit different characteristics, behaviors, and exposure, the modeling approaches are applied separately to a combined VRU and pedestrian-only datasets to understand their individual safety concerns.
Our study reveals that while both VRUs share some common surrogates that are significant predictors of confirmed conflicts, there are notable differences between them. 

The integration of machine learning techniques into our automated conflict detection framework has the potential to enhance the identification of complex patterns and relationships in the surrogate data. As a result, more reliable and robust conflict detection models can be developed, which could facilitate more effective safety interventions and reduce roadway crashes and fatalities. However, it is important to note that besides the automatically generated surrogates, the manual identification of confirmed conflicts also relies on subjective characteristics such as the level of recklessness displayed by road users, non-verbal communication, intersection features, and misjudgments. These subjective factors may not be accurately captured by current technology, which can limit the performance of the models.

\section{Acknowledgments}
The authors would like to thank the Pennsylvania Department of 
Transportation, Imperial Traffic Data Collection, LingaTech, Rybinski Engineering, Transoft Solutions for providing the data used in this analysis.

\bibliography{manuscript}

\end{document}